\documentclass{INTERSPEECH2023}

\interspeechcameraready

\usepackage{subcaption}
\usepackage{algorithm}
\usepackage{algpseudocode}

\usepackage{xcolor}
\usepackage{soul}

\newcommand{\hlc}[2][yellow]{{%
    \colorlet{foo}{#1}%
    \sethlcolor{foo}\hl{#2}}%
}

\usepackage{tipa}

\usepackage{stfloats} %

\title{Mai Ho`omāuna i ka `Ai: \\ Language Models Improve Automatic Speech Recognition in Hawaiian}
\name{Kaavya Chaparala$^1$, Guido Zarrella$^1$, 
Bruce Torres Fischer$^2$, Larry Kimura$^2$, 
Oiwi Parker Jones$^3$}
\address{
  $^1$The MITRE Corporation, USA\\
  $^2$University of Hawai`i at Hilo\\ %
  $^3$University of Oxford, UK}
\email{\{kchaparala,jzarrella\}@mitre.org, 
\{bruce42,larrykim\}@hawaii.edu, 
oiwi@robots.ox.ac.uk}

\begin{document}

\maketitle

\begin{abstract}
In this paper we address the challenge of improving Automatic Speech Recognition (ASR) for a low-resource language, Hawaiian, by incorporating large amounts of independent text data into an ASR foundation model, Whisper. To do this, we train an external language model (LM) on $\sim$1.5M words of Hawaiian text. We then use the LM to rescore Whisper and compute word error rates (WERs) on a manually curated test set of labeled Hawaiian data. As a baseline, we use Whisper without an external LM. 
Experimental results reveal a small but significant improvement in WER when ASR outputs are rescored with a Hawaiian LM. The results support leveraging all available data in the development of ASR systems for underrepresented languages.

\end{abstract}

\section{Introduction}\label{sec:1_intro}
    \begin{figure*}[!b] %
    \centering
    \begin{subfigure}[b]{0.45\textwidth}
        \includegraphics[width=\textwidth]{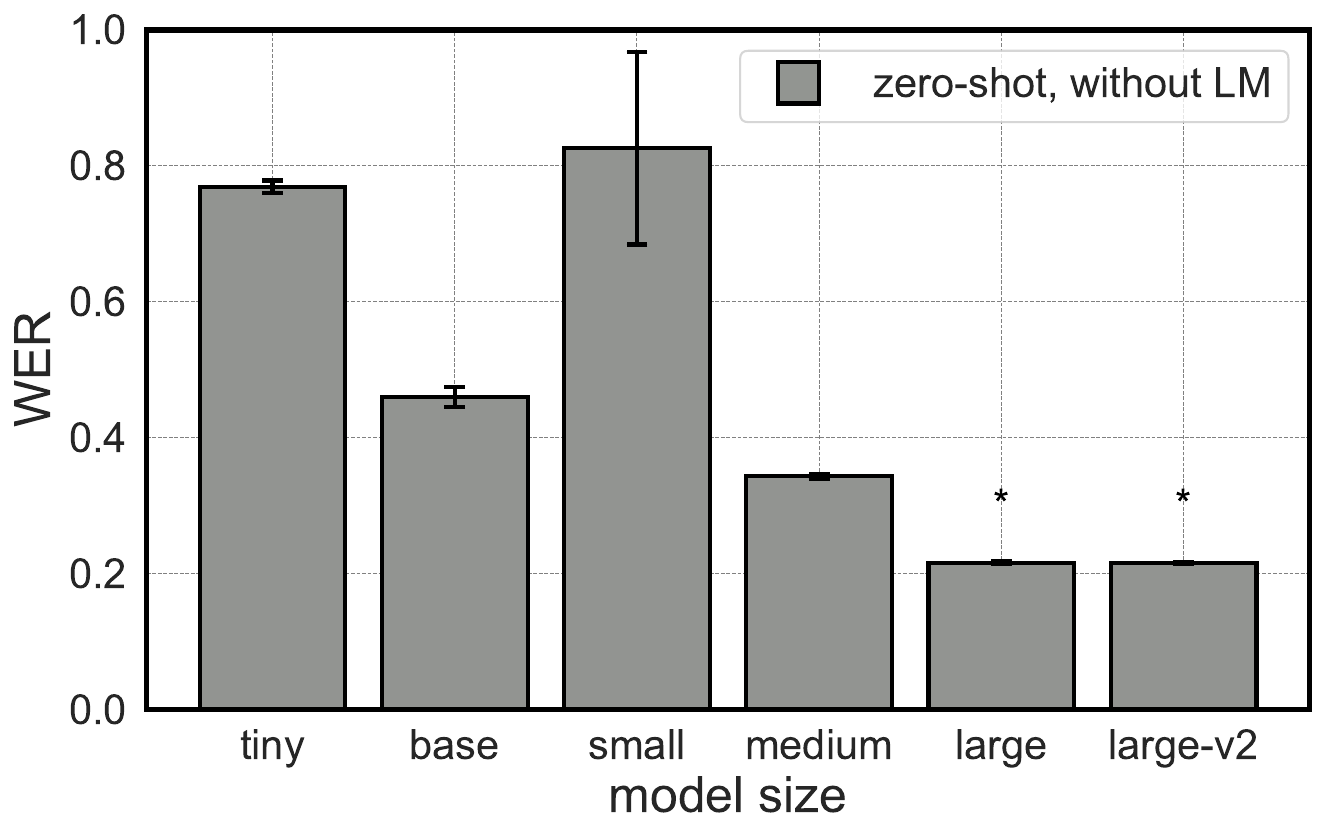}
    \end{subfigure}
    \hfill
    \begin{subfigure}[b]{0.45\textwidth}
        \includegraphics[width=\textwidth]{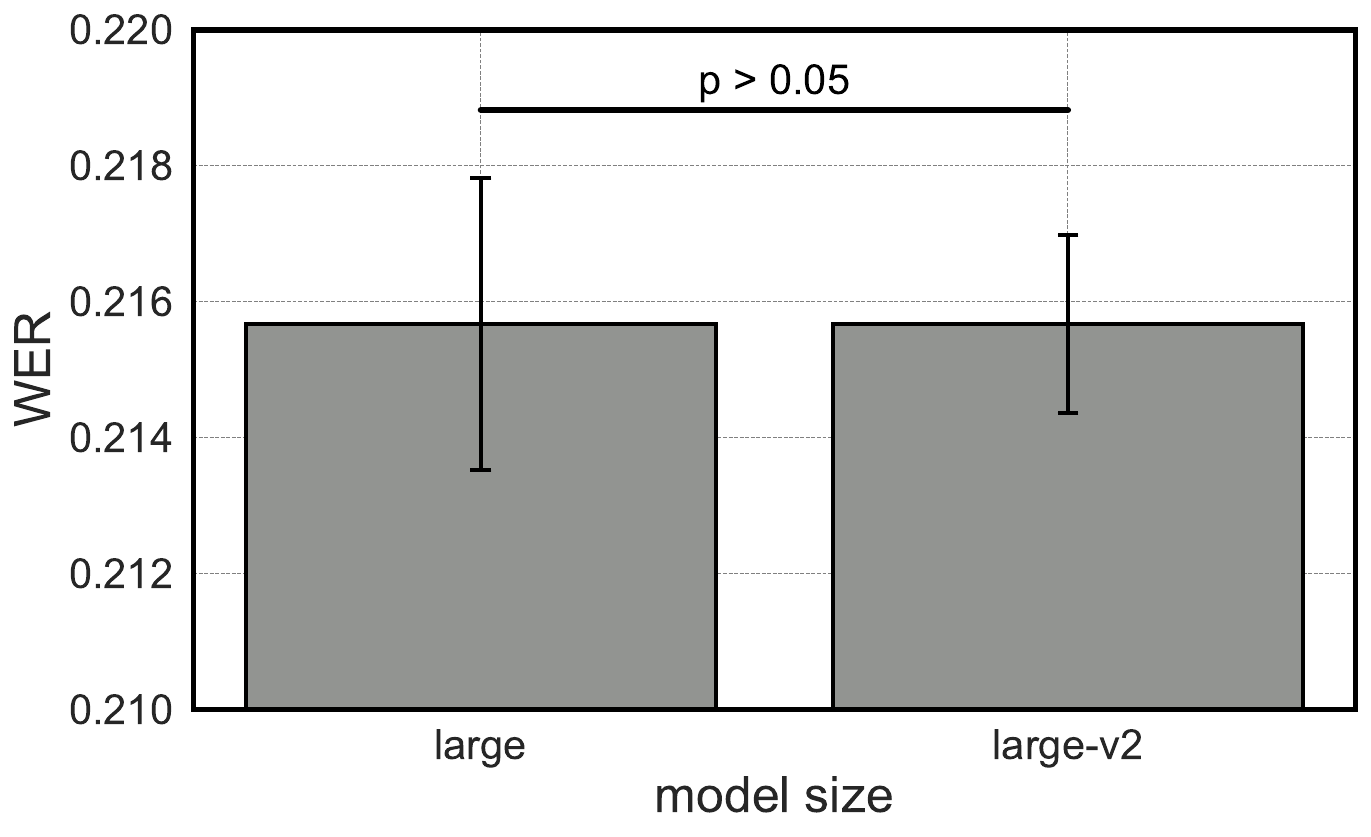}
    \end{subfigure}
    \caption{\textbf{Large ASR models produce the lowest word error rates (WERs) for Hawaiian test data.} Left panel: We compared six Whisper models on Hawaiian using zero-shot transfer without a Hawaiian language model (LM), as a baseline for comparing ASR models with LMs. Asterisks indicate the best models, \texttt{large} and \texttt{large-v2}. Right panel: No statistical difference in WER was observed between \texttt{large} and \texttt{large-v2} ($t_{3.309} = 0.002, p = 0.999$, Welch's t-test). Error bars show standard error of the mean.}
    \label{fig1:zero_shot_model_comparison}
\end{figure*}
Advances in Automatic Speech Recognition (ASR) have predominantly benefited a few global languages. 
To date, this has left speakers of low-resource languages, such as Hawaiian, at a disadvantage. 
High-quality Hawaiian ASR would support language preservation and revitalization efforts. 
In Hawaiian there is a traditional saying -- \textit{Mai ho`om\={a}una i ka `ai} -- which translates literally as `Don't waste food' \cite{pukui1983}. It can also be applied in the context of ASR to mean `Leave nothing to waste' or even `Use all the data you have'.
In this paper, we explore the use of text corpora as a potentially underutilized form of data to improve low-resource ASR systems. 
We do this by comparing %
a foundation model, Whisper \cite{radford2022}, both with and without the use of a relatively large amount of independent Hawaiian text. To leverage these data, we use an external language model (LM).

Unlike other early ASR foundation models (e.g.~wav2vec 2.0 \cite{baevski2020wav2vec2} and XLS-R \cite{babu2022xlsr}), 
Whisper is explicitly intended for \emph{zero-shot transfer} \cite{radford2022}. This means that the model is trained on a large and diverse set of data to generalize well on new ASR tasks without fine-tuning. 
The first contribution of this paper is to quantify Whisper's ability to perform zero-shot transfer in Hawaiian. We quantify performance using word error rates (WERs) calculated on a manually curated test set of labeled data (i.e.~Hawaiian audio--text pairs).

Prior work has explored how to improve foundation models like Whisper by incorporating independent large language models (LLMs) for English \cite{sun2023}. %
To the best of our knowledge, no similar work has been reported for low-resources languages like Hawaiian. 
Therefore, the second contribution of this paper is to evaluate the utility of incorporating independent Hawaiian text data into Whisper. 
We do this by replicating a state-of-the-art Hawaiian LM \cite{shillingford+parkerjones2018, parkerjones+shillingford2018} which we train on a relatively large corpus of modern Hawaiian text ($\sim$1.5M words). 
We then compare a baseline zero-shot Whisper model against a zero-shot Whisper model that has been rescored \cite{jurafsky+martin2024} using the Hawaiian LM.

    \begin{figure*}[ht]
    \centering
    \begin{subfigure}[b]{0.32\textwidth}
        \includegraphics[width=\textwidth]{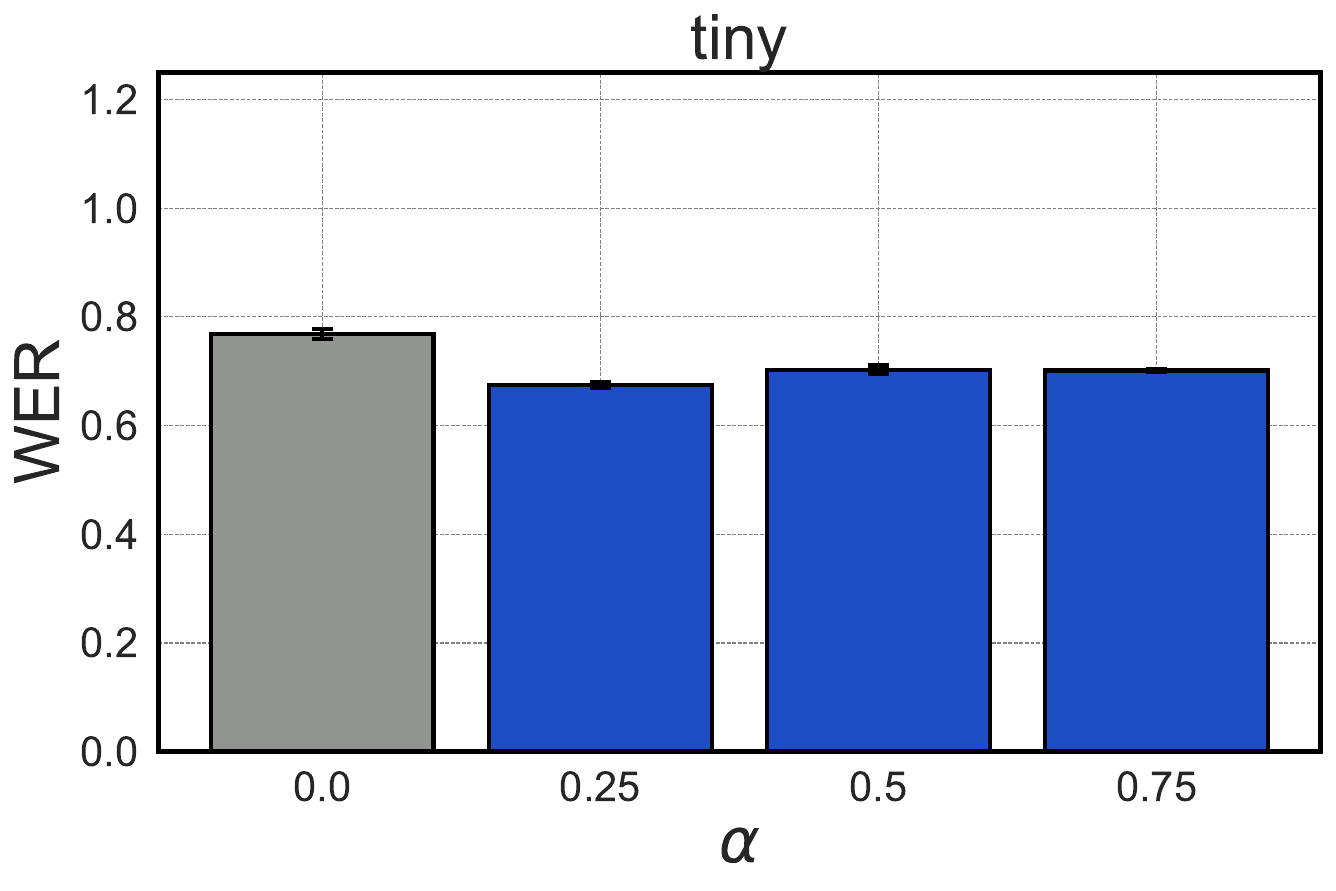}
    \end{subfigure}
    \hfill
    \begin{subfigure}[b]{0.32\textwidth}
        \includegraphics[width=\textwidth]{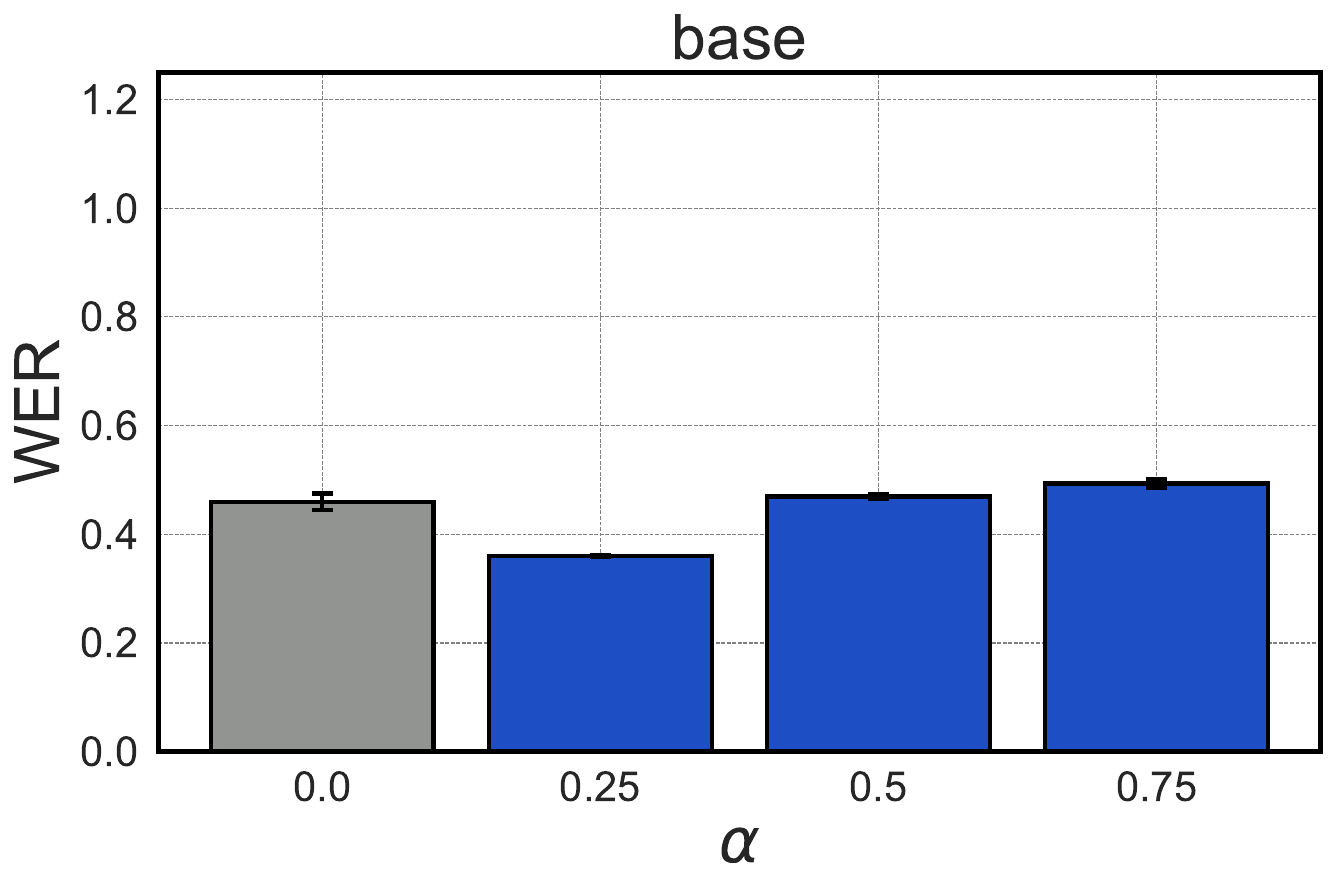}
    \end{subfigure}
    \hfill
    \begin{subfigure}[b]{0.32\textwidth}
        \includegraphics[width=\textwidth]{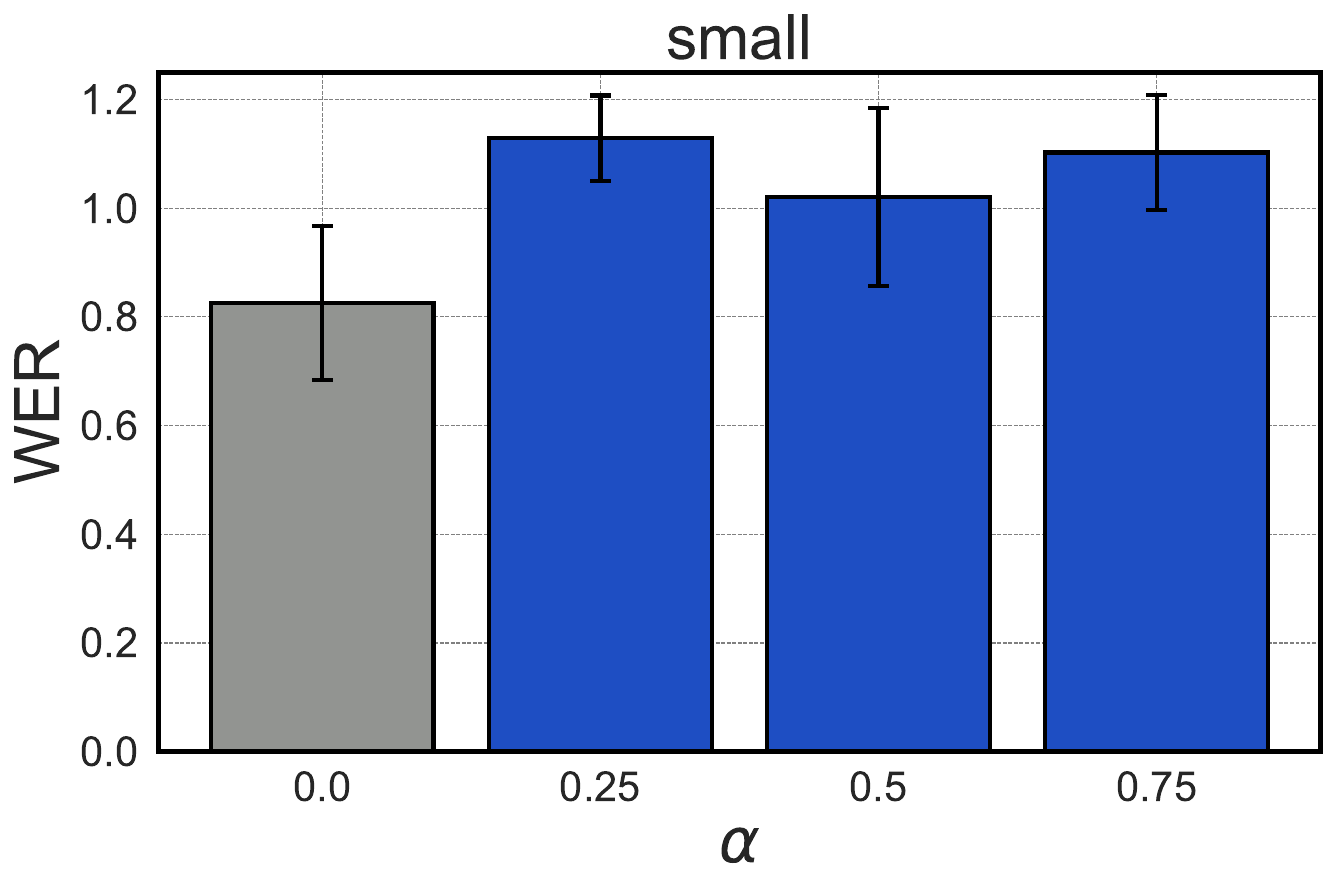}
    \end{subfigure}

    \begin{subfigure}[b]{0.32\textwidth}
        \includegraphics[width=\textwidth]{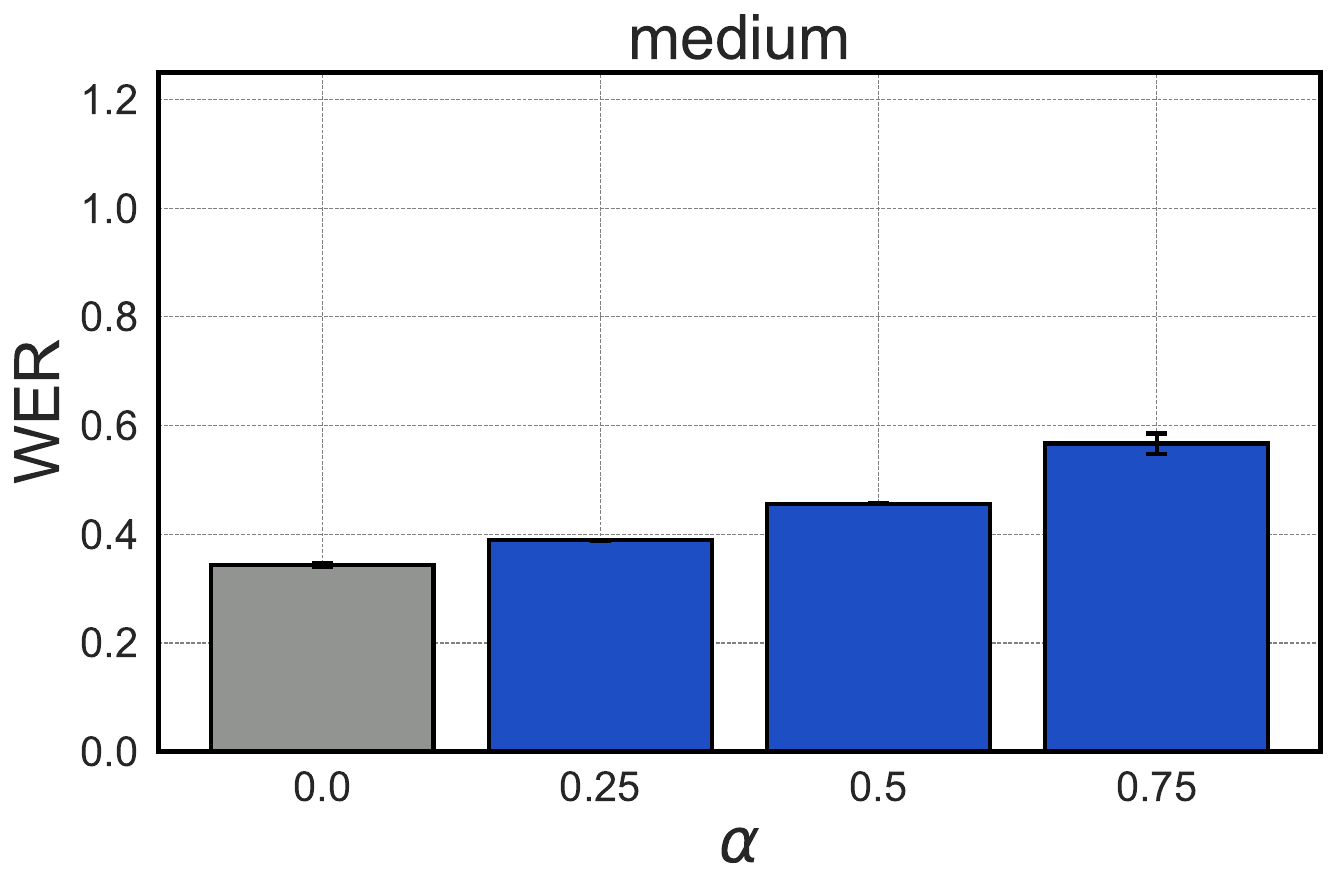}
    \end{subfigure}
    \hfill
    \begin{subfigure}[b]{0.32\textwidth}
        \includegraphics[width=\textwidth]{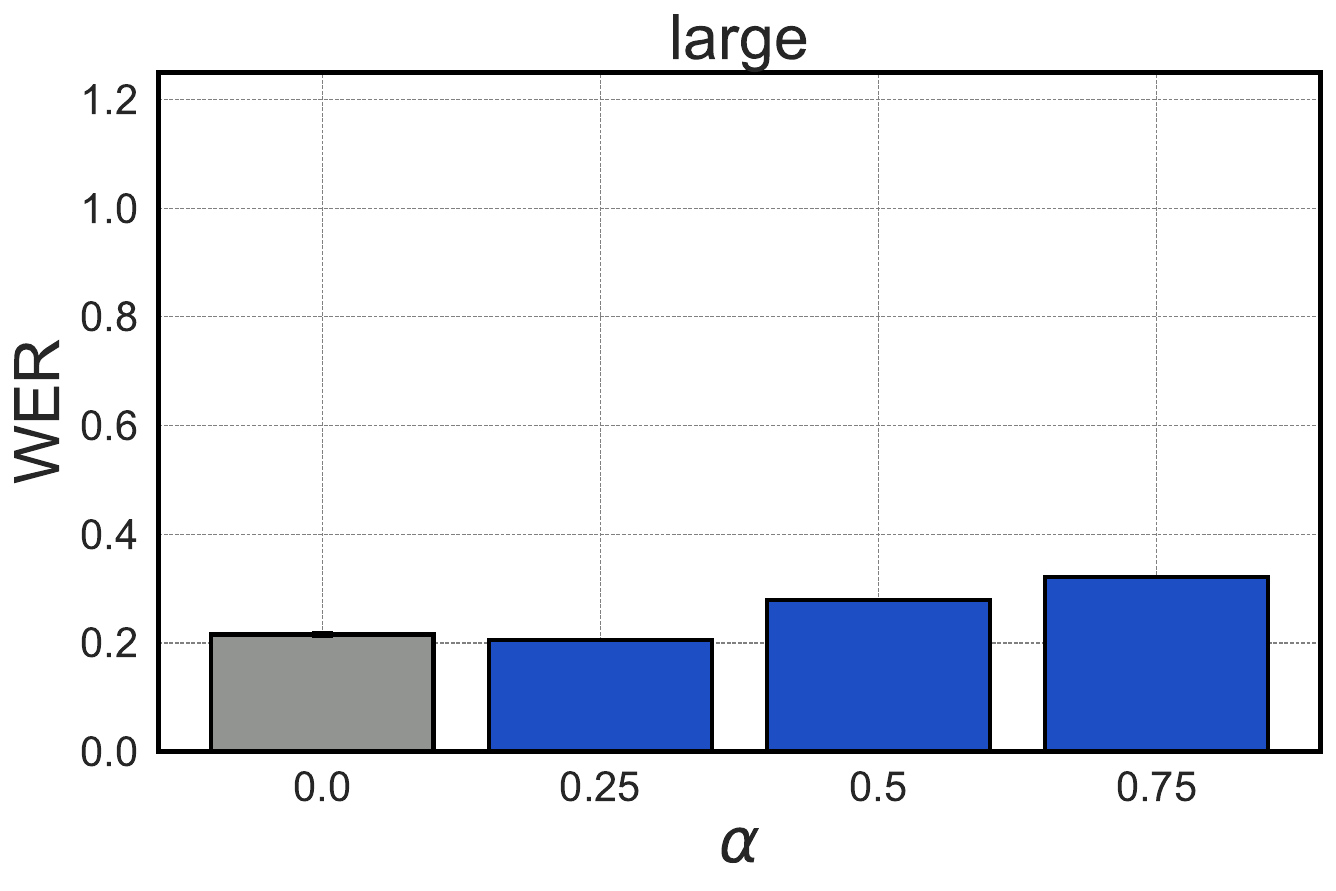}
    \end{subfigure}
    \hfill
    \begin{subfigure}[b]{0.32\textwidth}
        \includegraphics[width=\textwidth]{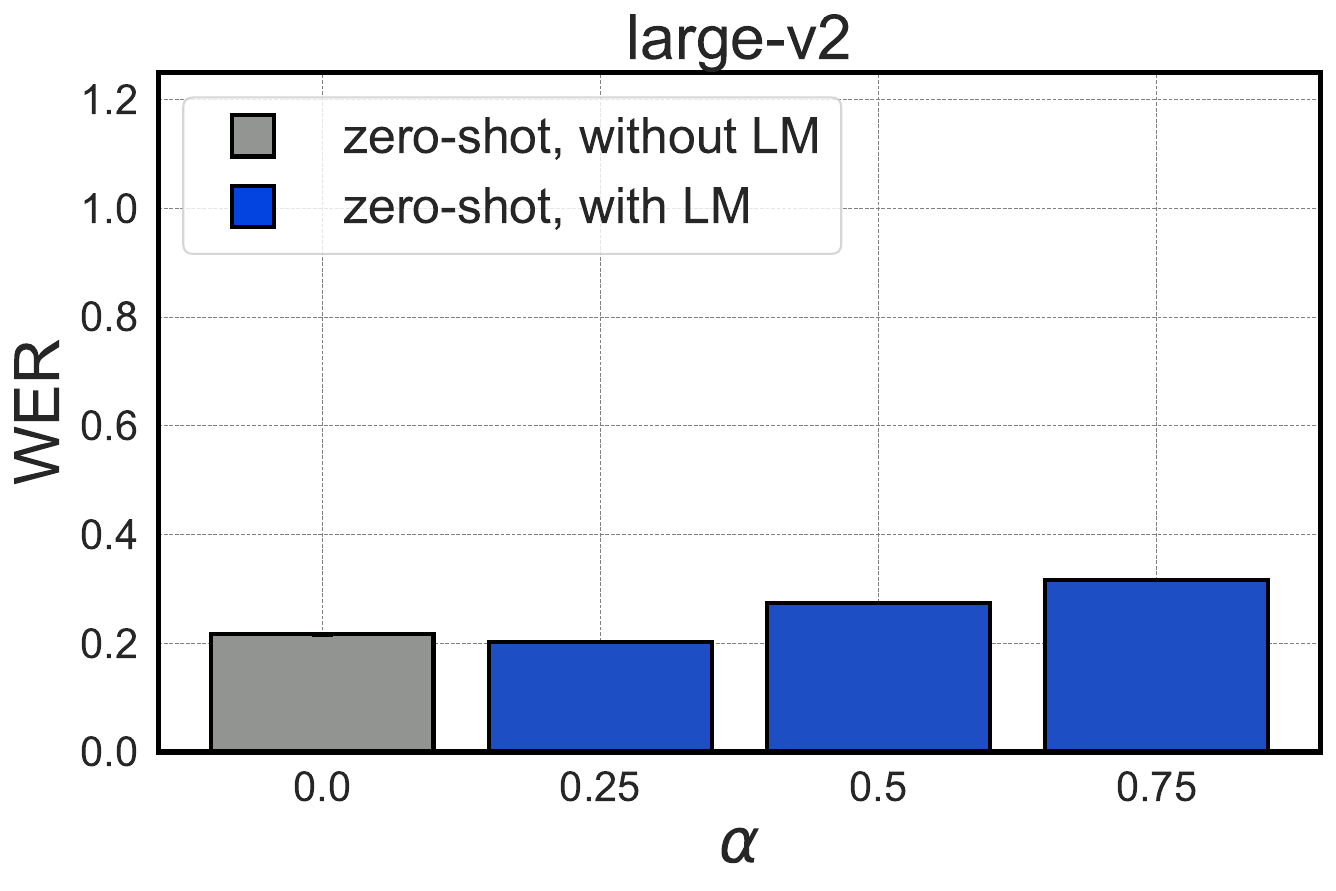}
    \end{subfigure}

    \caption{\textbf{An overview of zero-shot transfer in Hawaiian using Whisper models without LMs (gray bars) and with LMs (blue bars)}. The $\alpha$ values in the x-axes show the weighting of Hawaiian LMs against the ASR predictions (see text for details). Each panel presents results for different Whisper model. The difference in the best models (e.g.~\texttt{large-v2}) is small and hard to appreciate with the y-limits fixed at this scale. For \texttt{large-v2} results (bottom-right panel) with re-scaled y-limits, please see Figure \ref{fig2:rescore_large-v2}.}
    \label{fig3:rescore_all}
\end{figure*}

\section{Methods}\label{sec:3_methods}

\subsection{Hawaiian LMs}
Here we replicate the model architecture and training parameters from prior work on Hawaiian LMs \cite{shillingford+parkerjones2018, parkerjones+shillingford2018}, using the same dataset and data splits. 
Concretely, the training set, which we used for training LMs, alternately consists of 45,769 lines, 1,547,831 words, or 7,573,569 characters. The validation set, which we used for evaluating the validation loss and perplexity, consists of 888 lines, 26,607 words, or 129,487 characters. 
We did not use the test set (888 lines, 30,181 words, 146,124 characters) for this project. 
All modern Hawaiian texts (e.g.~\cite{hiiaka}) were used with permission.%

For the model architecture we followed prior work \cite{shillingford+parkerjones2018, parkerjones+shillingford2018} in implementing a character-level RNN \cite{sutskever2011rnns, graves2013rnns} which consisted of three layers of LSTMs \cite{hochreiter1997lstm} each with 200 features in their hidden states. The output of the final LSTM layer was passed to a linear layer that mapped to the set of output characters. 
We applied Dropout \cite{srivastava2014dropout} with a probability of 0.2 after the input and after each hidden layer. 
For the loss, we used cross entropy. 
LMs were trained using Adam \cite{kingma2014adam} for 10,000 epochs with an initial learning rate of 0.001, a batch size of 256, a clip value of 1, and a maximum sequence length of 100 characters. %
During training, we checkpointed the LM that produced the lowest perplexity on the validation set. In the final LM which we used for rescoring, the validation perplexity was 2.024. This is, incidentally, better than the reported state-of-the-art validation perplexity of 2.65 \cite{parkerjones+shillingford2018}, a difference that we attribute to our checkpointing strategy. %

For rescoring, the LM was used to compute the log probability of strings. This computation involved prepending a start-of-sequence (SOS) token to the string. The LM needs this to transition into the first character in the string. Since the LM character set did not include an explicit SOS token, we opted to use a whitespace character instead. This decision was based on the rationale that, within the context of the LM, the probability distribution following a whitespace character should closely approximate the distribution of initial characters in words.

Formally, the log probability of observing a character $y_i$, given the LM and preceding sequence of characters, is denoted as $\log P_{\text{LM}}(y_i | y_{1:i-1})$, where $y_{1:i-1}$ represents the sequence of characters from the start of the string up to, but not including, $y_i$. 
Consequently, the log probability of an entire string $Y = y_1, y_2, \dots, y_n$ is computed as the sum of its individual characters, given by their respective preceding character sequences: $\log P_{\text{LM}}(Y) = \sum_{i=1}^n \log P_{\text{LM}}(y_i | y_{1:i-1})$. 

\begin{figure}[ht]
    \includegraphics[width=.45\textwidth]{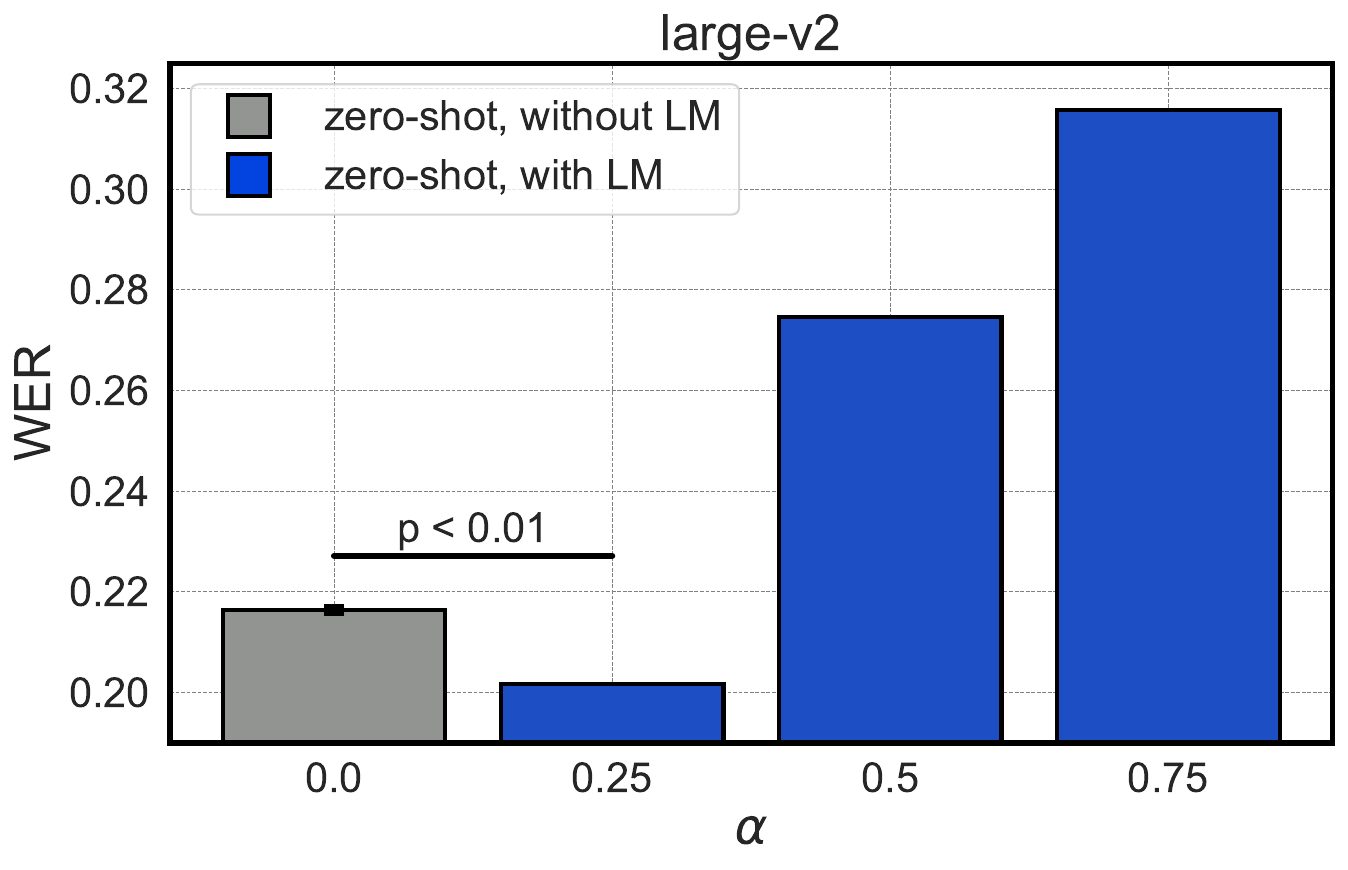}
    \caption{\textbf{Rescoring with a Hawaiian LM provides a small but significant improvement on the zero-shot Whisper baseline.} Rescoring results for \texttt{large-v2}. The $\alpha$ values weight the contribution of the LM. $\alpha=0$ means no contribution of the LM (baseline model). Other values add increasing weight to the LM. The best WER was found at $\alpha=0.25$ where we observe a small but significant improvement on the baseline model ($t_2 = 19.498, p = 0.003$, one-sample t-test).}
    \label{fig2:rescore_large-v2}
\end{figure}

\begin{figure*}[!thb] %
    \centering
    \begin{subfigure}[b]{0.32\textwidth}
        \includegraphics[width=\textwidth]{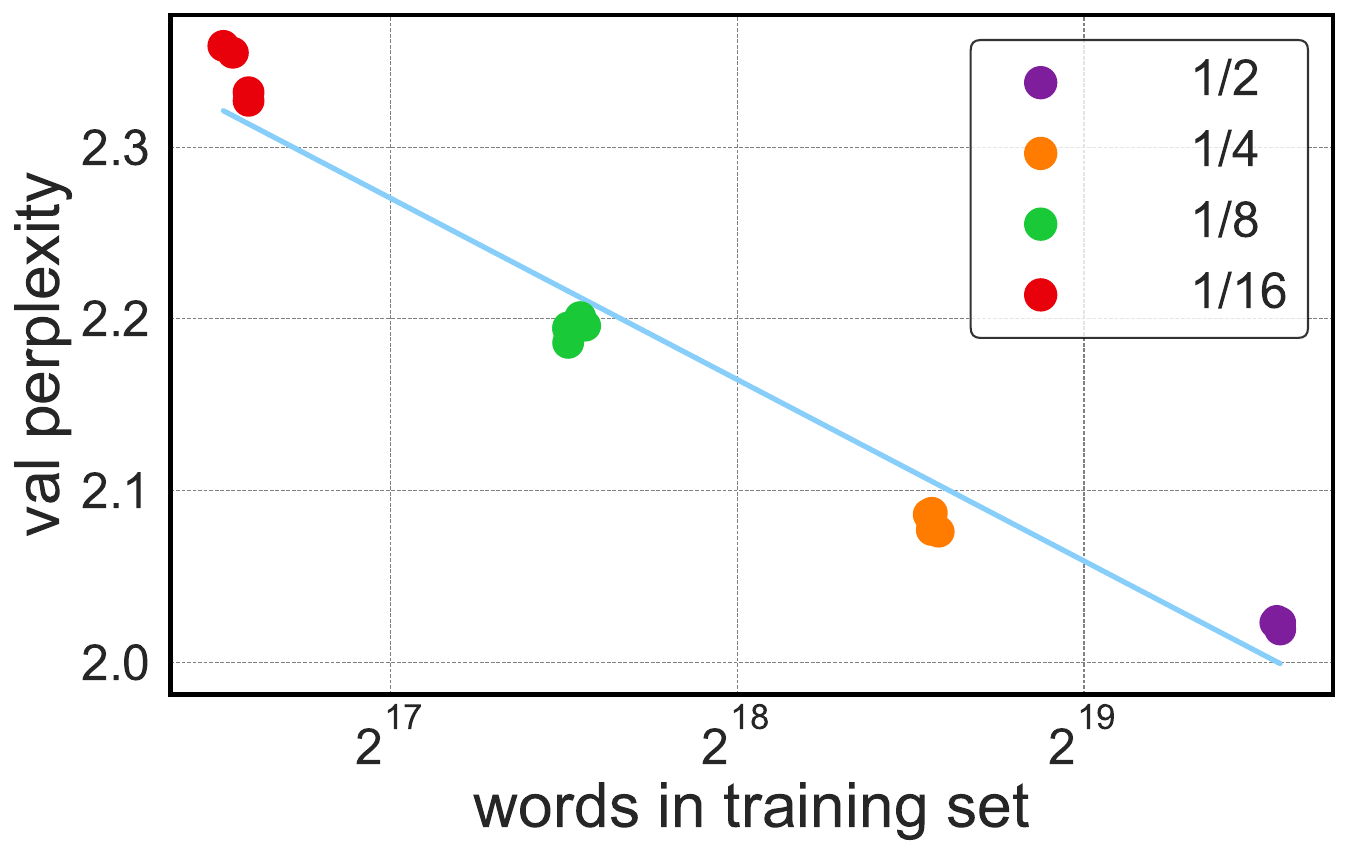}
    \end{subfigure}
    \hfill
    \begin{subfigure}[b]{0.32\textwidth}
        \includegraphics[width=\textwidth]{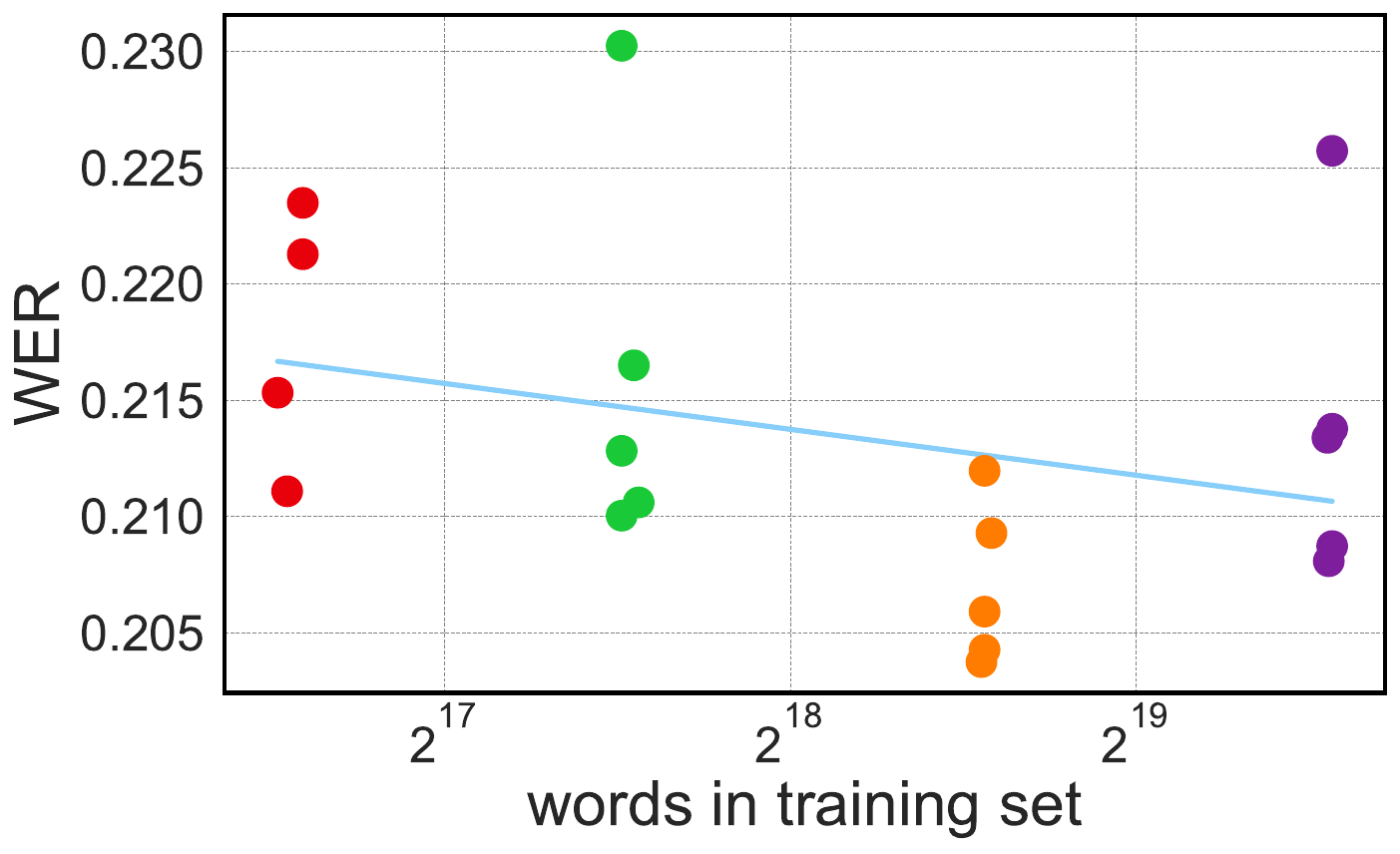}
    \end{subfigure}
    \hfill
    \begin{subfigure}[b]{0.32\textwidth}
        \includegraphics[width=\textwidth]{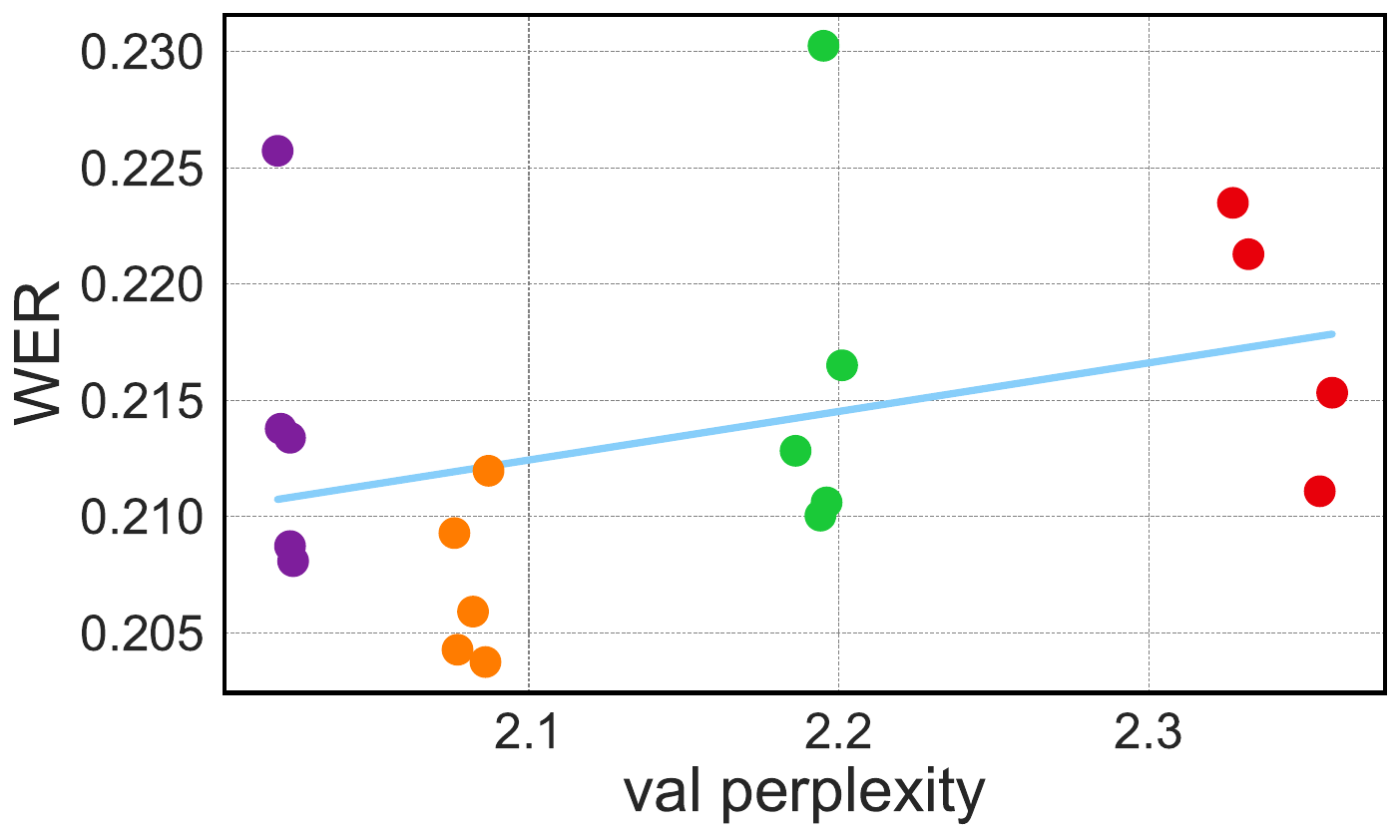}
    \end{subfigure}
    \caption{\textbf{Posthoc exploration on the amount of training text, LM validation perplexity, and Whisper WER.} Hawaiian LMs were trained on decreasing fractions of data: 1/2 (purple), 1/4 (orange), 1/8 (green), and 1/16 (red). 
    See text for details. 
    }
    \label{fig4:posthoc}
\end{figure*}

\subsection{Rescoring Whisper}
Encoder--decoder models, such as Whisper \cite{radford2022}, are essentially conditional language models that learn an implicit language model over the domain of output tokens in their training data. 
Critically, Whisper's implicit language model is not tuned to Hawaiian, as no Hawaiian text was included during training.  We note that some Hawaiian audio was paired with English text in Whisper's $X$$\to$\texttt{en} task \cite{radford2022}. But Whisper was not trained on any Hawaiian audio--text pairs, or indeed any Hawaiian text.
We therefore propose to incorporate an explicit Hawaiian LM into Whisper by using rescoring. 
In rescoring \cite{jurafsky+martin2024}, the score used during beam search combines the ASR model's log probability and the log probability of the external LM. 
To see this, let the ASR model's log probability be $\log P_{\text{ASR}}(Y | X)$, where $Y$ is the output text and $X$ is the input audio. 
We combine this with the log probability of the external LM, $\log P_{\text{LM}}(Y)$, as follows:
\begin{equation}
    \text{score} = \alpha \log P_{\text{LM}}(Y) + (1 - \alpha) \log P_{\text{ASR}}(Y | X)
\end{equation}
Here $\alpha$ is a hyperparameter that balances the influence of the log probabilities of the LM and ASR models. 
As expressed, $\alpha=0$ means no contribution from the LM. 
We explore values of $\alpha \in \{ 0, 0.25, 0.5, 0.75 \}$. We exclude $\alpha = 1$ as this would mean no influence from the ASR. 

In the Hugging Face implementation that we used,\footnote{\url{https://huggingface.co/docs/transformers/en/model_doc/whisper}} %
Whisper employs a multi-stage decoding process. It begins with beam search, utilizing a default beam size of 5. However, if the output does not meet predefined compression ratio and log-probability thresholds, the implementation iteratively resorts to greedy search with temperatures varying between 0.2 to 1.0. These heuristics aim to ensure that the final hypothesis avoids common error modes for encoder--decoder models, such as repeating substrings in the output or failing to generate an end-of-text token.
We highlight these heuristics as they affect whether model outputs are deterministic or stochastic. This in turn affected our choice of statistical test when comparing different Whisper models (one sample vs independent samples t-tests).

\subsection{ASR Test Set}
The evaluation of WERs required a labeled ASR test set for Hawaiian. To this end, we utilized a subset of the publicly available \textit{Ka Leo Hawai`i} (KLH) dataset, being careful to account for transcription and segmentation errors within the data. First, we randomly selected 100 audio--text pairs from the dataset. We then filtered the pairs manually, discarding any that, after text normalization, resulted in empty strings (e.g.~from audio segments annotating only laughter). We also discarded pairs that showed discrepancies between audio and text (e.g.~truncated recordings). Such exclusions were crucial to avoid artificially inflating the WER through misalignments. After filtering 57 pairs remained, comprising 1,120 words and a total audio duration of 7 minutes and 35.336 seconds. This constituted our final test set. As Whisper does not process audio files longer than 30 seconds, it is important to note that the longest audio file in the ASR test set was only 26.593 seconds in duration.

\section{Experiment Results}\label{sec:4_experiments}
\subsection{Which Whisper model transfers best to Hawaiian?}
Whisper comes in different sizes: \texttt{tiny}, \texttt{base}, \texttt{small}, \texttt{medium}, and \texttt{large}. 
Multiple \texttt{large} models were also available when we ran the experiments: \texttt{large} and \texttt{large-v2}. 
We compared all of these using zero-shot transfer on the ASR test for Hawaiian, repeating the evaluation three times for each model to estimate the variability in WER results. 
The best results (Figure \ref{fig1:zero_shot_model_comparison}, left panel) were obtained for the \texttt{large} and \texttt{large-v2} models. 
Focusing on the best two models (Figure \ref{fig1:zero_shot_model_comparison}, right panel), we find no significant difference between \texttt{large} and \texttt{large-v2} in terms of zero-shot WER ($t_{3.309}=0.002, p=0.999$, Welch's t-test). 
As either model could be used, we focus below on rescoring \texttt{large-v2}.

\subsection{Does rescoring improve the best Whisper model?}
Figure \ref{fig3:rescore_all} summarizes the rescoring results for all model sizes. 
It shows that rescoring improves the WER in some cases (e.g.~\texttt{base}) but not all (e.g.~\texttt{small, medium}). %
Turning to the \texttt{large-v2} model, highlighted in Figure \ref{fig2:rescore_large-v2}, we find that rescoring with the Hawaiian LM produces a small absolute improvement ($\sim$1--2\%) in WER when compared to the zero-shot baseline at $\alpha=0.25$. 
This improvement is, however, significant ($t_2 = 19.498, p = 0.003$, one-sample t-test). 
To get a sense of what predictions from the model look like, we can sample a few examples at random. 
Predicted transcripts from the best rescored zero-shot model are compared against ground truth transcripts in Table \ref{tab:asr_examples}.

\subsection{Does it matter how much text the LM is trained on?}
In a final set of posthoc experiments, we re-trained a series of Hawaiian language models on fractions of the full dataset used in the main results. 
Starting with the full training set (45,769 lines), we randomly sampled subsets of training text with varying fractions of lines: 1/2, 1/4, 1/8, and 1/16. For each fraction of lines, we created five randomly sub-sampled training sets. These were then used to train 20 new LMs (4 fractions $\times$ 5 repeats) using the same training parameters as in the main analysis. These 20 LMs were then used to rescore the \texttt{large-v2}  Whisper model. 
In Figure \ref{fig4:posthoc}, we show scatter plots that compare the number of words in the sub-sampled training set, the trained LM's validation perplexity, and the \texttt{large-v2} Whisper model's rescored WER. 

We observe a negative correlation between the number of words in the training set and the LM's validation perplexity ($r = -0.98, p = 8.67 \times 10^{-13}$, Pearson's correlation, Bonferroni corrected, 3 comparisons). More text for training corresponds to smaller validation perplexity (Figure \ref{fig4:posthoc}, left panel). 

Correlations involving WER did not reach significance, apparently because of large variance in the WER results. 
However, linear regression models suggest trends. 
For example, although not statistically significant ($r = -0.308, p = 0.14$ uncorrected), the trend is for WER to decrease as training data increases (Figure \ref{fig4:posthoc}, middle panel). 
Similarly, although not statistically significant ($r = 0.352, p = 0.14$ uncorrected), WER appears to increase as validation perplexity increases (Figure \ref{fig4:posthoc}, right panel). 
Or, equivalently, in this last case, WER appears to decrease as validation perplexity decreases. 
Again, we note that correlations involving WERs did not reach statistical significance. 
With a larger compute budget, it might be possible to resolve the variance in WERs. 
In the present work, we used about 1,000 GPU hours (279 training LMs and 730 evaluating Whisper models).

    \begin{table*}[t]
\centering
\caption{\textbf{Example ASR predictions and ground truth sentences suggest directions for future work.} 
Here we show three predictions from the best Whisper model (\texttt{large-v2}) rescored with the best LM (1/1 training data). 
To facilitate the discussion in the text, character-level edits are colored for \hlc[red]{deletion}, \hlc[yellow]{substitution}, and \hlc[green]{insertion}. 
Differences in capitalization and punctuation are ignored, as the text was normalized before evaluation.
Elsewhere in the paper, WERs were computed at the word-level. 
See text for discussion.}
\label{tab:asr_examples}
\begin{tabular}{r|p{6cm}|p{6cm}}
\hline
\textbf{ID} & \textbf{ASR Prediction} & \textbf{Ground Truth} \\ \hline
1 & O L\hlc[yellow]{a}haina ke kapikala hiko o Hawai`i. & \hlc[green]{`}O L\hlc[yellow]{ā}haina ke kapikala \hlc[green]{ka}hiko o Hawai`i. \\
2 & Makemake wau e ike i nā wai e\hlc[red]{ }hā. & Makemake wau e \hlc[green]{`}ike i Nā Wai \hlc[green]{`}Ehā. \\
3 & A\hlc[yellow]{i}, e hele māka\hlc[yellow]{ }i\hlc[red]{ }ka\hlc[yellow]{ }in\hlc[yellow]{o} k\hlc[yellow]{a}ua. & \hlc[green]{`}A\hlc[yellow]{e}. E hele māka\hlc[yellow]{`}ika\hlc[yellow]{`}i\hlc[green]{ }n\hlc[yellow]{ō} k\hlc[yellow]{ā}ua. \\
\hline
\end{tabular}
\end{table*}
\section{Discussion}\label{sec:5_discussion}
There are dozens of hours of labeled Hawaiian audio but millions of pages of Hawaiian newspaper text available right now. %
We therefore investigated, in this paper, how the use of text data might improve Hawaiian ASR. 
We also evaluated the capacity for a foundation model, Whisper \cite{radford2022}, to accurately transcribe Hawaiian audio without having seen any labeled Hawaiian data. 
Out of the box, we found the largest Whisper models (\texttt{large} and \texttt{large-v2}) could transcribe Hawaiian audio with WERs of about 22\% (Figure \ref{fig3:rescore_all}). 
By incorporating a Hawaiian LM, the \texttt{large-v2} model achieves a WER of about 20\% (Figure \ref{fig2:rescore_large-v2}). 
Ultimately, the difference in WER is only about 1--2\% but the improvement is consistent across repeated evaluations ($p < 0.01$). 
For low-resource languages like Hawaiian, which could use better ASR to accelerate language preservation and revitalization efforts, any statistically significant improvement in performance is welcome. But there is more work to do.

To suggest ways of improving on these results, let us consider where they fail. 
In Table \ref{tab:asr_examples}, we see three ASR predictions from the Whisper \texttt{large-v2} model together with corresponding ground truth transcriptions. 
Character-level errors have been highlighted to indicate deletions, substitutions, and insertions when mapping from the prediction to the ground truth transcript. 
A number of intriguing patterns emerge. 
For example, we see that the model frequently fails to capture phonemic glottal stops /\textipa{P}/ (written in Hawaiian with the $\langle$`$\rangle$ symbol, which in Hawaiian is called an \textit{`okina} /\textipa{P}okina/). 
The phoneme /\textipa{P}/ can be realized in Hawaiian speech with full closure, modal voice, or creak \cite{parkerjones2018, davidson2021, davidson+parkerjones2023}. 
Acoustically, these are relatively quiet sounds. 
In a few cases (the glottal stops in \textit{māka\hlc[yellow]{`}ika\hlc[yellow]{`}i}), we see that the model tries to capture these with spaces. 
This may relate to a bias for English phonology to insert a glottal stop before a word-initial vowel. 
This would also account for the missing glottal stop on \textit{\hlc[green]{`}Ae}. 
That said, not all of the extra spaces added by the model (e.g.~\textit{e\hlc[red]{ }hā}, \textit{i\hlc[red]{ }ka}) correspond to missing glottal stops and perhaps show that the model has not properly learned to segment Hawaiian words. 
Or, since \textit{e}, \textit{hā}, \textit{i}, and \textit{ka},  are all attested Hawaiian words \cite{pukui+elbert1986}, it may be more accurate to say the model has not learned to predict the correct word given the context. 

Another interesting failure mode is the treatment of vowels. 
In many cases, the model fails to distinguish phonemically short and long vowels (e.g.~\textit{L\hlc[yellow]{ā}haina} /\textipa{la:.hai.na}/, \textit{n\hlc[yellow]{ō}} /\textipa{no:}/, \textit{k\hlc[yellow]{ā}ua} /\textipa{ka:u.a}/).\footnote{The ground truth transcription for \textit{L{ā}haina} may be disputed as the place name is also written and pronounced without the long vowel: \textit{Lahaina}. In this case, the model prediction would be correct.} 
Noticeably, the model also gets the vowel wrong on the diphthong \textit{`A\hlc[yellow]{e}}. 
We note that the quality of a vowel in Hawaiian can differ whether in a monophthong or diphthong \cite{parkerjones2018, kettig2021}. 
All of these errors suggest that the model does poorly on sounds that diverge from English, which makes sense given that Whisper is optimized for English. 
We hypothesize that these errors may be corrected by fine-tuning Whisper on sufficient amounts of labeled Hawaiian data. 
Serious efforts are currently underway within the Hawaiian community to gather and organize a much larger collection of labeled Hawaiian data and we are actively working to leverage all of it to improve Hawaiian ASR. 

Following the principle of using all the data at hand, the Hawaiian community has access to a lot more unlabeled text, and unlabeled audio. 
On the text side, one might train a better LM, or even LLM, for Hawaiian (e.g.~using Transformers \cite{vaswani2017transformers}) while scaling up the size of the training set. 
One might also advance text normalization methods \cite{shillingford+parkerjones2018, parkerjones+shillingford2018}. These could then be used to modernize a larger amount of the 19th century texts that were written in an older missionary orthography which excludes letters for Hawaiian sounds that do not occur in English \cite{schutz1994}. 
Finally, two obvious ways to incorporate unlabeled audio would be to use self-supervised learning \cite{oord2018representation, schneider2019wav2vec, chung2019unsupervised, baevski2020vqwav2vec, baevski2020wav2vec2, chung2021w2vbert, hsu2021hubert, hsu2021hubertjournal} and pseudo-labeling \cite{lee2013pseudo, synnaeve2020end, park2020improved, xu2020iterative, likhomanenko2021slimipl, berrebbi2023continuousPL}.

What can other low-resource language communities learn from this study? 
We conjecture that many languages are like Hawaiian in having much less labeled data than unlabeled data (e.g.~text corpora). 
Part of our motivation for including the posthoc experiments, which varied the amount of text data used to train an LM, was to provide a guide for languages with less text data. 
Nonetheless, when it comes to leveraging models like Whisper, Hawaiian may not be representative. %
For example, other languages may not be included in Whisper's training data. 
After all, Whisper was trained on 338 hours of unlabeled audio from Hawaiian as well as 1381 hours of unlabeled audio from M\={a}ori, a closely related Polynesian language \cite{radford2022}. 
The Hawaiian alphabet is also very similar to English, with only a few modifications for glottal stops and long vowels. 
Ultimately, how well our approach works for other low-resource languages is an empirical question. 
But that may be enough. For languages that have few options, we offer hope: it is worth trying, as Hawaiian tradition teaches, to use all the data you have.

\section{Acknowledgments}

Mahalo iā Ka`iu Kimura ma Ka Haka `Ula O Ke`elikōlani no ke kāko`o a iā Puakea Nogelmeier ma Awaiaulu no ka `ae `ana iā mākou e ho`ohana i ka puke `o \textit{Hi`iakaikapoliopele} \cite{hiiaka}. %
We are grateful to Shuo Chen, Brendan Shillingford, Michael and Caroline Running Wolf, Keoni Mahelona, Miles Thompson, Jason Lewis, Lisa Davidson, Tatiana Likhomanenko, and Dan Jurafsky for discussing various aspects of this work. For constructive feedback we also thank audiences at the University of Oxford and at two Indigenous AI Workshops (at ICML 2023 and at NeurIPS 2023). We acknowledge NVIDIA for the generous contribution of GPUs used in this research. We also thank our funders for their support: OPJ is supported by the MRC (MR/X00757X/1), Royal Society (RG$\backslash$R1$\backslash$241267), NSF (2314493), NFRF (NFRFT-2022-00241), and SSHRC (895-2023-1022).
BTF and LK are supported by the NSF (1664070) and NEH (PD-255910-17). 
Additional support was provided by 
MITRE's Independent Research \& Development Program. Approved for Public Release; Distribution Unlimited. Public Release Case Number 24-0733. %

\bibliographystyle{IEEEtran}
\bibliography{mybib}

\end{document}